\RequirePackage{fix-cm}
\documentclass[twocolumn,compsoc,tikz]{cvm}
\pdfoutput=1
\usepackage{tikz}
\usetikzlibrary{matrix,chains,positioning,decorations.pathreplacing,arrows}
\usepackage{import}
\usetikzlibrary{quotes,arrows.meta}
\usetikzlibrary{positioning}

\usepackage{Ball}
\usepackage{Box}
\usepackage{Block}
\usepackage{RightBandedBox}

\usetikzlibrary{positioning}
\usetikzlibrary{3d} 
\usepackage{graphicx,wrapfig,lipsum}
\usepackage{adjustbox}
\usepackage{hyperref}
\graphicspath{{figures/},{figures/cvm/}}
\usepackage{etoolbox}

\def\ManuscriptInfo{Manuscript received: 2020-02-20; accepted: 2020-04-24.}

\def\PaperTitle{iSeeBetter: Spatio-temporal video super-resolution using recurrent generative back-projection networks}

\def\AuthorNames{Aman Chadha$^1$(\rm{\Letter}), John Britto$^2$, and M. Mani Roja$^3$}

\def\AuthorAdress{$1\quad $ &Department of Computer Science, Stanford University, 450 Serra Mall, Stanford, CA 94305, USA. E-mail: aman@amanchadha.com; amanc@stanford.edu (\rm{\Letter}).\\
    $2\quad $ & Department of Computer Science, University of Massachusetts Amherst, Amherst, MA 01003, USA. E-mail: jnadar@umass.edu.\\
   $3\quad $ & Department of Electronics and Telecommunication Engineering, University of Mumbai, Mumbai, Maharashtra 400032, India. E-mail: maniroja@tsec.edu\\
}

\def\firstheader{DOI 10.1007/s41095-xxx-xxxx-x\hfill Vol. x, No. x, month year, xx--xx\\[5mm]~Article Type (Research/Review)}

\headevenname{{Aman Chadha \etal}}
\robustify{\caption}
\robustify{\path}
\begin{document}

\MakePageStyle

\MakeAbstract{Recently, learning-based models have enhanced the performance of single-image super-resolution (SISR). However, applying SISR successively to each video frame leads to a lack of temporal coherency. Convolutional neural networks (CNNs) outperform traditional approaches in terms of image quality metrics such as peak signal to noise ratio (PSNR) and structural similarity (SSIM). However, generative adversarial networks (GANs) offer a competitive advantage by being able to mitigate the issue of a lack of finer texture details, usually seen with CNNs when super-resolving at large upscaling factors. We present iSeeBetter, a novel GAN-based spatio-temporal approach to video super-resolution (VSR) that renders temporally consistent super-resolution videos. iSeeBetter extracts spatial and temporal information from the current and neighboring frames using the concept of recurrent back-projection networks as its generator. Furthermore, to improve the ``naturality'' of the super-resolved image while eliminating artifacts seen with traditional algorithms, we utilize the discriminator from super-resolution generative adversarial network (SRGAN). Although mean squared error (MSE) as a primary loss-minimization objective improves PSNR/SSIM, these metrics may not capture fine details in the image resulting in  misrepresentation of perceptual quality. To address this, we use a four-fold (MSE, perceptual, adversarial, and total-variation (TV)) loss function. Our results demonstrate that iSeeBetter offers superior VSR fidelity and surpasses state-of-the-art performance.}
\MakeKeywords{super resolution; video upscaling; frame recurrence; optical flow; generative adversarial networks; convolutional neural networks}
\section{Introduction}\label{sec:introduction}

The goal of super-resolution (SR) is to enhance a low resolution (LR) image to a higher resolution (HR) image by filling in missing fine-grained details in the LR image. The domain of SR research can be divided into three main areas: single image SR (SISR) \cite{dong2015image, haris2018deep, haris2017inception, kim2016accurate}, multi image SR (MISR) \cite{faramarzi2013unified, garcia2012super} and video SR (VSR) \cite{caballero2017real, tao2017detail, sajjadi2018frame, haris2019recurrent, jo2018deep}.

Consider an LR video source which consists of a sequence of LR video frames $LR_{t-n}$, ..., $LR_{t}$, ..., $LR_{t+n}$, where we super-resolve a target frame $LR_{t}$. 
The idea behind SISR is to super-resolve $LR_{t}$ by utilizing spatial information inherent in the frame, independently of other frames in the video sequence. However, this technique fails to exploit the temporal details inherent in a video sequence resulting in temporal incoherence. 
MISR seeks to address just that -- it utilizes the missing details available from the neighboring frames $LR_{t-n}$, ..., $LR_{t}$, ..., $LR_{t+n}$ and fuses them for super-resolving $LR_{t}$. After spatially aligning frames, missing details are extracted by separating differences between the aligned frames from missing details observed only in one or some of the frames. However, in MISR, the alignment of the frames is done without any concern for temporal smoothness, 
which is in stark contrast to VSR where the frames are typically aligned in temporal smooth order.

Traditional VSR methods upscale based on a single degradation model (usually bicubic interpolation) followed by reconstruction. This is sub-optimal and adds computational complexity \cite{shi2016real}. Recently, learning-based models that utilize convolutional neural networks (CNNs) have outperformed traditional approaches in terms of widely-accepted image reconstruction metrics such as peak signal to noise ratio (PSNR) and structural similarity (SSIM). 

In some recent VSR methods that utilize CNNs, frames are concatenated \cite{jo2018deep} or fed into recurrent neural networks (RNNs) \cite{huang2015bidirectional} in temporal order, without explicit alignment. In other methods, the frames are aligned explicitly, using motion cues between temporal frames with the alignment modules \cite{caballero2017real, liu2017robust, tao2017detail, sajjadi2018frame}. The latter set of methods generally render temporally smoother results compared to the methods with no explicit spatial alignment \cite{liao2015video, huang2015bidirectional}. However, these VSR methods suffer from a number of problems. In the frame-concatenation approach \cite{caballero2017real, liu2017robust, jo2018deep}, many frames are processed simultaneously in the network, resulting in significantly higher network training times. With methods that use RNNs \cite{sajjadi2018frame, tao2017detail, huang2015bidirectional}, modeling both subtle and significant changes simultaneously (e.g., slow and quick motions of foreground objects) is a challenging task even if long short-term memory units (LSTMs) are deployed, which are  designed for maintaining long-term temporal dependencies \cite{gers1999learning}. 
A crucial aspect of an effective VSR system is the ability to handle motion sequences, which are often integral components of videos \cite{caballero2017real, makansi2017end}.

\begin{figure}[!b]
    \vspace {-4mm}
    \centering
    \includegraphics[width=0.8\linewidth]{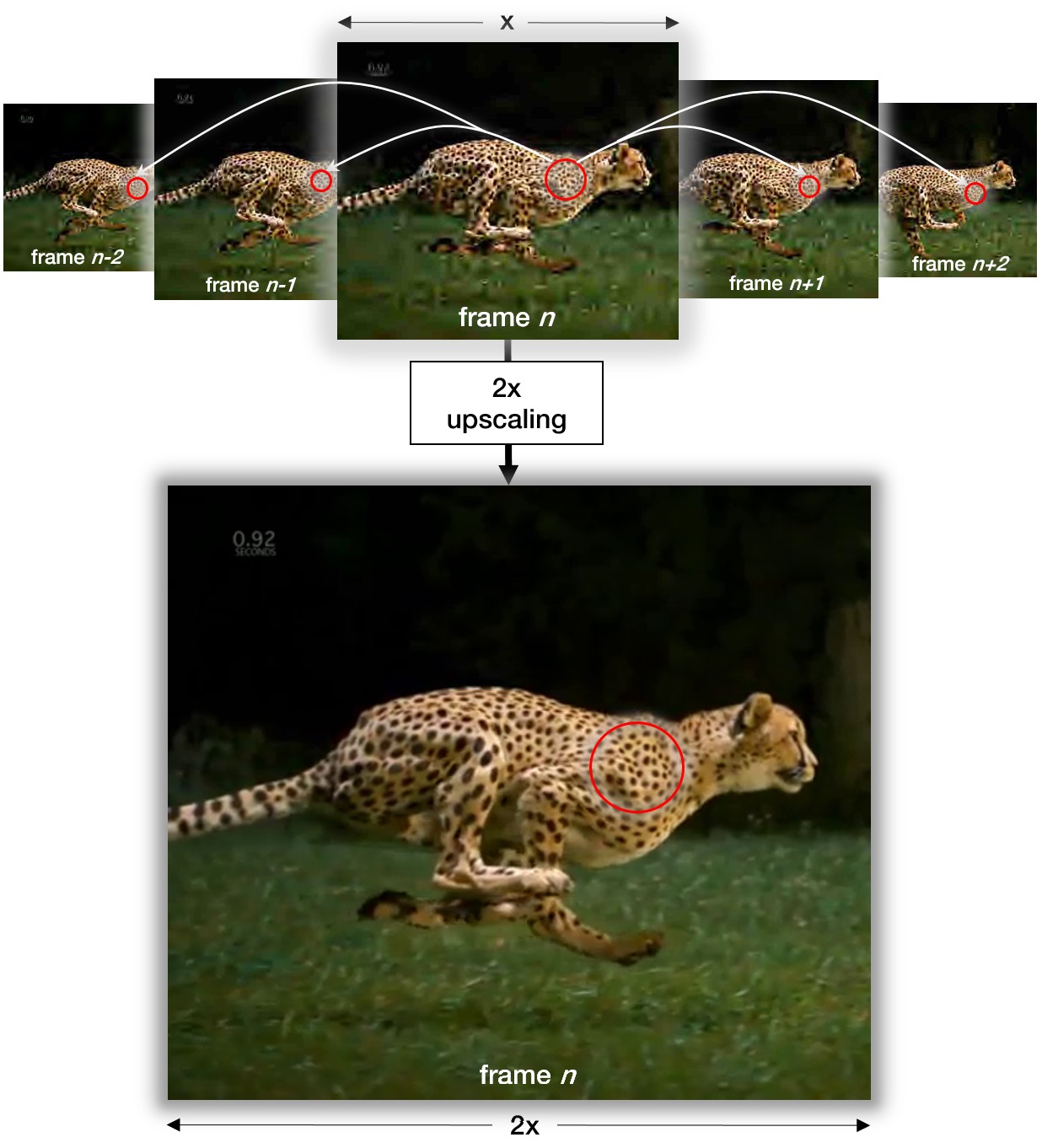} \vspace {-1mm}
    \caption{Temporal relationships between adjacent frames.}
    \label{fig:1}
\end{figure}

The proposed method, iSeeBetter, is inspired by recurrent back-projection networks (RBPNs) \cite{haris2019recurrent} which utilize ``back-projection'' as their underpinning approach, originally introduced in \cite{irani1991improving, irani1993motion} for MISR. 
The basic concept behind back-projection is to iteratively calculate residual images as reconstruction error between a target image and a set of neighboring images. The residuals are then back-projected to the target image for improving super-resolution accuracy. The multiple residuals enable representation of subtle and significant differences between the target frame and its adjacent frames, thus exploiting temporal relationships between adjacent frames as shown in Fig. \ref{fig:1}. Deep back-projection networks (DBPNs) \cite{haris2018deep} use back-projection to perform SISR using learning-based methods by estimating the output frame $SR_{t}$ using the corresponding $LR_{t}$ frame. To this end, DBPN produces a high-resolution feature map that is iteratively refined through multiple up- and down-sampling layers. RBPN offers superior results by combining the benefits of the original MISR back-projection approach with DBPN. Specifically, RBPN uses the idea of iteratively refining HR feature maps from DBPN, but extracts missing details using neighboring video frames like the original back-projection technique \cite{irani1991improving, irani1993motion}. 
This results in superior SR accuracy.

To mitigate the issue of a lack of finer texture details when super-resolving at large upscaling factors that is usually seen with CNNs \cite{ledig2017photo}, iSeeBetter utilizes GANs with a loss function that weighs adversarial loss, perceptual loss \cite{ledig2017photo}, mean square error (MSE)-based loss and total-variation (TV) loss \cite{frsrgan}. Our approach combines the merits of RBPN and SRGAN \cite{ledig2017photo} -- it is based on RBPN as its generator and is complemented by SRGAN's discriminator architecture, which is trained to differentiate between super-resolved images and original photo-realistic images.
Blending these techniques yields iSeeBetter, a state-of-the-art system that is able to recover precise photo-realistic textures and motion-based scenes from heavily down-sampled videos. 

Our contributions include the following key innovations. 

\textbf{Combining the state-of-the-art in SR}: We propose a model that leverages two superior SR techniques -- (i) RBPN, which is based on the idea of integrating SISR and MISR in a unified VSR framework using back-projection and, (ii) SRGAN, which is a framework capable of inferring photo-realistic natural images. 
RBPN enables iSeeBetter to extract details from neighboring frames, complemented by the generator-discriminator architecture in GANs which pushes iSeeBetter to generate more realistic and appealing frames while eliminating artifacts seen with traditional algorithms \cite{wang2020deep}. iSeeBetter thus yields more than the sum of the benefits of RBPN and SRGAN.

\textbf{``Optimizing'' the loss function}: 
Pixel-wise loss functions such as L1 loss, used in RBPN \cite{haris2019recurrent}, struggle to handle the uncertainty inherent in recovering lost high-frequency details such as complex textures that commonly exist in many videos. 
Minimizing MSE encourages finding pixel-wise averages of plausible solutions that are typically overly-smooth and thus have poor perceptual quality \cite{mathieu2015deep, johnson2016perceptual, dosovitskiy2016generating, bruna2016super-resolution}. To address this, we adopt a four-fold (MSE, perceptual, adversarial, and TV) loss function for superior results. 
Similar to SRGAN \cite{ledig2017photo}, we utilize a loss function that optimizes perceptual quality by minimizing adversarial loss and MSE loss. Adversarial loss helps improve the ``naturality'' associated with the output image using the discriminator. On the other hand, MSE loss focuses on optimizing perceptual similarity instead of similarity in pixel space. Furthermore, we use a de-noising loss function called TV loss \cite{aly2005image}.
We carried out experiments comparing L1 loss with our four-fold loss and found significant improvements with the latter (cf. Section \ref{sec:evaluation}). 

\textbf{Extended evaluation protocol}: To evaluate iSeeBetter, we used standard datasets: Vimeo90K \cite{xue2019video}, Vid4 \cite{liu2011bayesian} and SPMCS \cite{tao2017detail}. 
Since Vid4 and SPMCS lack significant motion sequences, we included Vimeo90K, a dataset containing various types of motion. This enabled us to conduct a more holistic evaluation of the strengths and weaknesses of iSeeBetter. 
To make iSeeBetter more robust and enable it to handle real-world videos, we expanded the spectrum of data diversity and wrote scripts to collect additional data from YouTube. As a result, we augmented our dataset to about 170,000 clips.

\textbf{User-friendly infrastructure}: 
We built several useful tools to download and structure datasets, visualize temporal profiles of intermediate blocks and the output, and run predefined benchmark sequences on a trained model to be able to iterate on different models quickly. In addition, we built a video-to-frames tool to directly input videos to iSeeBetter, rather than frames. 
We also ensured our script infrastructure is flexible (such that it supports a myriad of options) and can be easily leveraged. The code and pre-trained models are available at \url{https://iseebetter.amanchadha.com}.

\begin{table*}[ht]
    \caption{Datasets used for training and evaluation}
    \label{tab:1}
    \setlength{\tabcolsep}{4.5pt}
\begin{adjustbox}{width=0.85\textwidth,center}    
    \begin{tabular}{ccccc}
    \toprule
Dataset & Resolution & \# of Clips & \# of Frames/Clip & \# of Frames\\\hline
 Vimeo90K & 448 $\times$ 256 & 13,100 & 7 & 91,701 \\
 SPMCS & 240 $\times$ 135 & 30 & 31 & 930\\
 Vid4 & (720 $\times$ 576 $\times$ 3) $\times$ 2, (704 $\times$ 576 $\times$ 3) $\times$ 2 & 4 & 41, 34, 49, 47 & 684\\
 \hline
 Augmented & 960 $\times$ 720 & 7,000 & ~110 & 77,000\\
 \hline
\textbf{Total} & - & \textbf{46,034} & - & \textbf{170,315} \\
        \bottomrule
    \end{tabular}
\end{adjustbox}    
\end{table*}

\section{Related work} \label{sec:Relatedwork}
Since the seminal work by Tsai on image registration \cite{tsai1984multiframe} two decades ago, many SR techniques based on various underlying principles have been proposed. Initial methods included spatial or frequency domain signal processing, statistical models and interpolation approaches \cite{yang2010image}. 
In this section, we focus our discussion on learning-based methods which have emerged as superior VSR techniques compared to traditional statistical methods.

\subsection{Deep SISR}
First introduced by SRCNN \cite{dong2015image}, deep SISR required a predefined up-sampling operator. 
Further improvements in this field include better up-sampling layers \cite{shi2016real}, residual learning \cite{tai2017image}, back-projection \cite{haris2018deep}, recursive layers \cite{kim2016deeply}, and progressive up-sampling \cite{lai2017deep}. 
A significant milestone in SR research was the introduction of a GAN-powered SR approach \cite{ledig2017photo}, which achieved state-of-the-art performance.

\subsection{Deep VSR}
Deep VSR can be divided into five types based on the approach to preserving temporal information.

\textbf{(a) Temporal Concatenation.} The most popular approach to retain temporal information in VSR is concatenating multiple frames \cite{kappeler2016video, caballero2017real, jo2018deep, liao2015video}. This approach can be seen as an extension of SISR to accept multiple input images. 
However, this approach fails to represent multiple motion regimes within a single input sequence since the input frames are simply concatenated together.

\textbf{(b) Temporal Aggregation.} To address the dynamic motion problem in VSR, \cite{liu2017robust} proposed multiple SR inferences which work on different motion regimes. The final layer aggregates the outputs of all branches to construct SR frame. However, this approach still concatenates many input frames, resulting in lengthy convergence during global optimization.

\textbf{(c) Recurrent Networks.} 
RNNs deal with temporal inputs and/or outputs and have been deployed in a myriad of applications ranging from video captioning \cite{johnson2016densecap, mao2014deep, yu2016video}, video summarization \cite{donahue2015long, venugopalan2014translating}, and VSR \cite{tao2017detail, huang2015bidirectional, sajjadi2018frame}. Two types of RNNs have been used for VSR. A many-to-one architecture is used in \cite{huang2015bidirectional, tao2017detail} where a sequence of LR frames is mapped to a single target HR frame. A many-to-many RNN has recently been used by \cite{sajjadi2018frame} where an optical flow network to accepts $LR_{t-1}$ and $LR_t$, which is fed to an SR network along with $LR_t$. This approach was first proposed by \cite{huang2015bidirectional} using bidirectional RNNs. However, the network has a small network capacity and has no frame alignment step. Further improvement is proposed by \cite{tao2017detail} using a motion compensation module and a ConvLSTM layer \cite{shiconvolutional}.

\textbf{(d) Optical Flow-Based Methods.} The above methods estimate a single HR frame by combining a batch of LR frames and are thus computationally expensive. They often result in unwanted flickering artifacts in the output frames \cite{frsrgan}. To address this, \cite{sajjadi2018frame} proposed a method that utilizes a network trained on estimating the optical flow along with the SR network. Optical flow methods allow estimation of the trajectories of moving objects, thereby assisting in VSR. \cite{kappeler2016video} warp video frames $LR_{t-1}$ and $LR_{t+1}$ onto $LR_{t}$ using the optical flow method of \cite{drulea2011total}, concatenate the three frames, and pass them through a CNN that produces the output frame $SR_{t+1}$. \cite{caballero2017real} follow the same approach but replace the optical flow model with a trainable motion compensation network.

\textbf{(e) Pre-Training then Fine-Tuning v/s End-to-End Training.} While most of the above-mentioned methods are end-to-end trainable, certain approaches first pre-train each component before fine-tuning the system as a whole in a final step \cite{caballero2017real, tao2017detail, liu2017robust}.

Our approach is a combination of (i) an RNN-based optical flow method that preserves spatio-temporal information in the current and adjacent frames as the generator and, (ii) a discriminator that is adept at ensuring the generated SR frame offers superior fidelity.

\section{Methods} \label{sec:methods}

\subsection{Datasets}\label{sec:datasets}

To train iSeeBetter, we amalgamated diverse datasets with differing video lengths, resolutions, motion sequences, and number of clips. Tab. \ref{tab:1} presents a summary of the datasets used. When training our model, we generated the corresponding LR frame for each HR input frame by performing 4$\times$ down-sampling using bicubic interpolation. We thus perform self-supervised learning by automatically generating the input-output pairs for training without any human intervention. 
To further extend our dataset, we wrote scripts to collect additional data from YouTube. The dataset was shuffled for training and testing. Our training/validation/test split was 80\%/10\%/10\%.

\subsection{Network architecture} \label{implementation}

\begin{figure*}[t]
    \vspace {-4mm}
    \centering
    \includegraphics[width=0.75\linewidth]{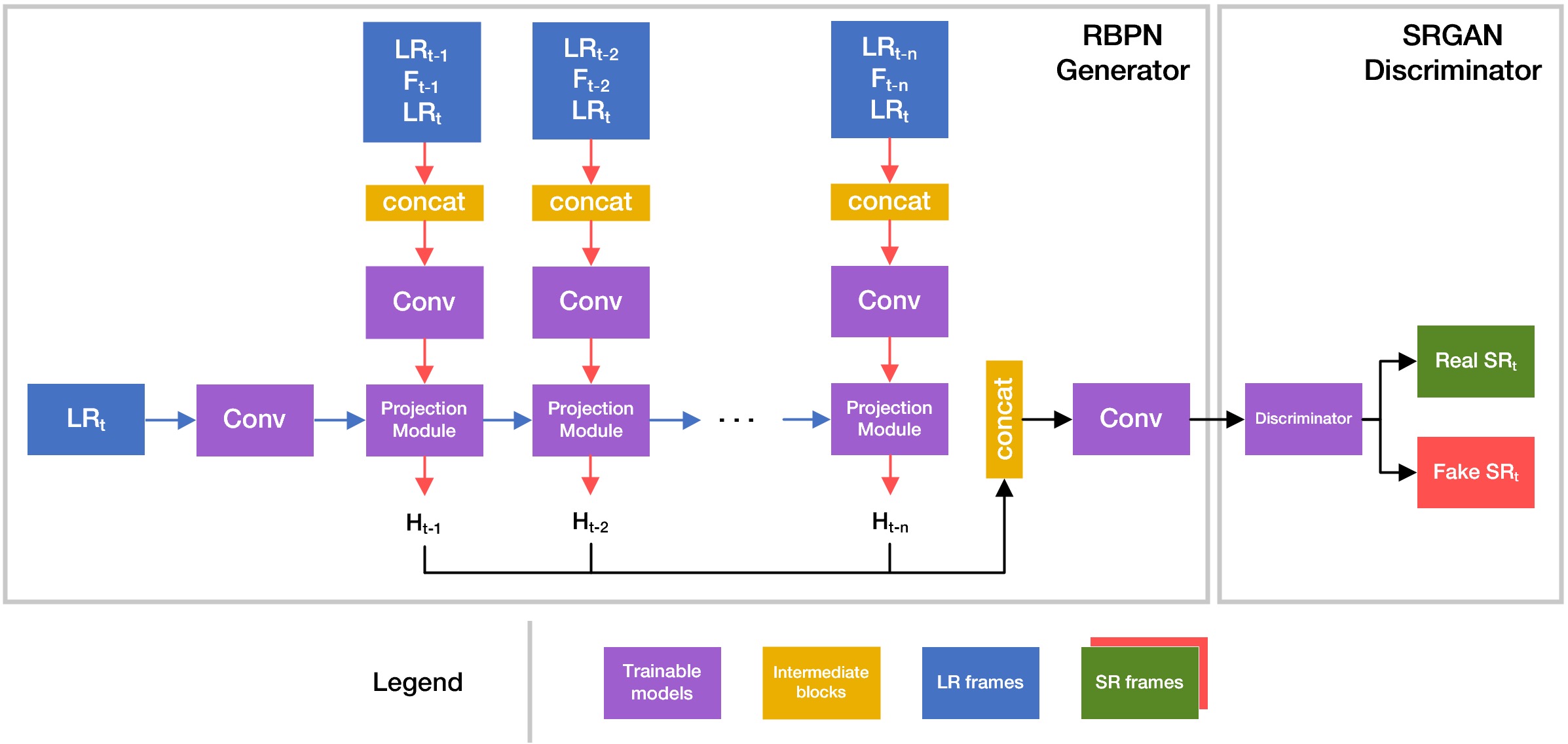} \vspace {-2mm}
    \caption{Overview of iSeeBetter}
    \label{fig:2}
\end{figure*}

\begin{figure*}
    \vspace {-4mm}
    \centering
    \includegraphics[width=0.95\linewidth]{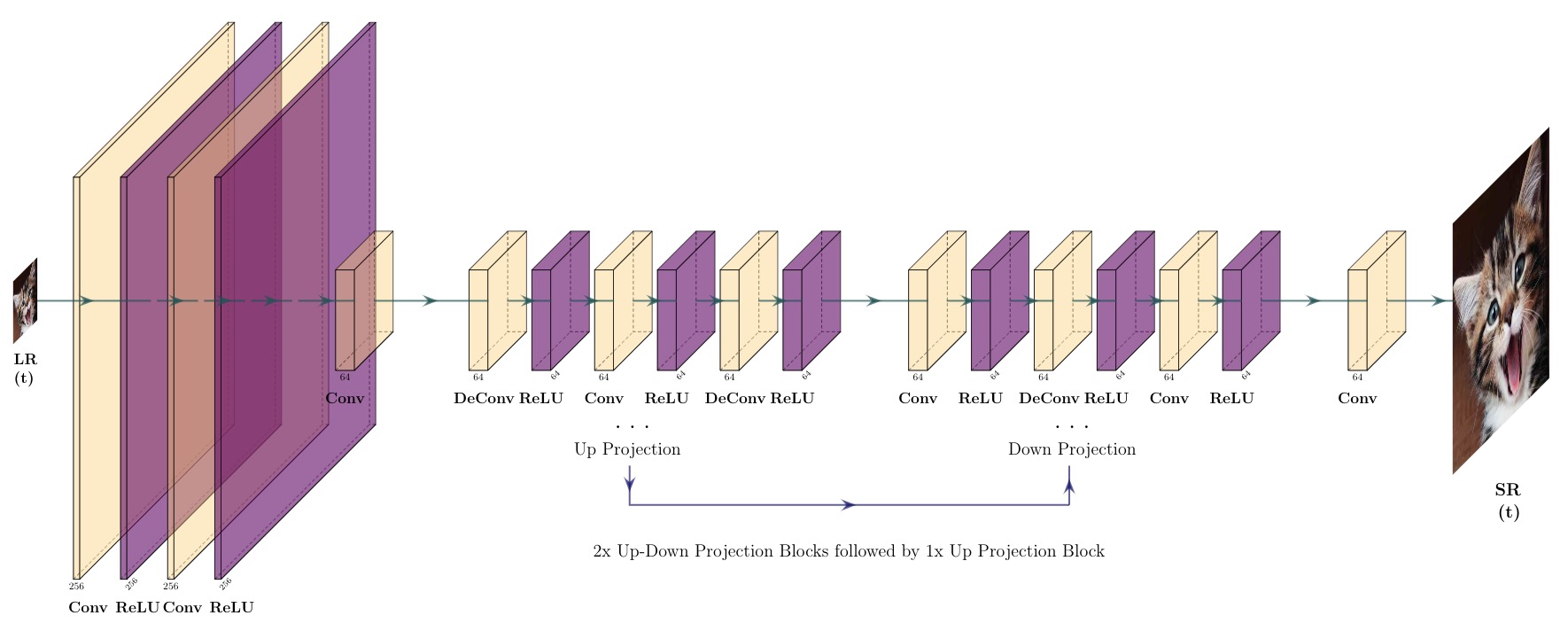} \vspace {-1mm}
    \renewcommand{\thefigure}{3}
\caption{DBPN \cite{haris2018deep} architecture for SISR, where we perform up-down-up sampling using 8 $\times$ 8 kernels with a stride of 4 and padding of 2. Similar to the ResNet architecture above, the DBPN network also uses Parametric ReLUs \cite{he2015delving} as its activation functions.}
    \label{fig:3}
\end{figure*}

\begin{figure*}
    \vspace {-4mm}
    \centering
    \includegraphics[width=0.95\linewidth]{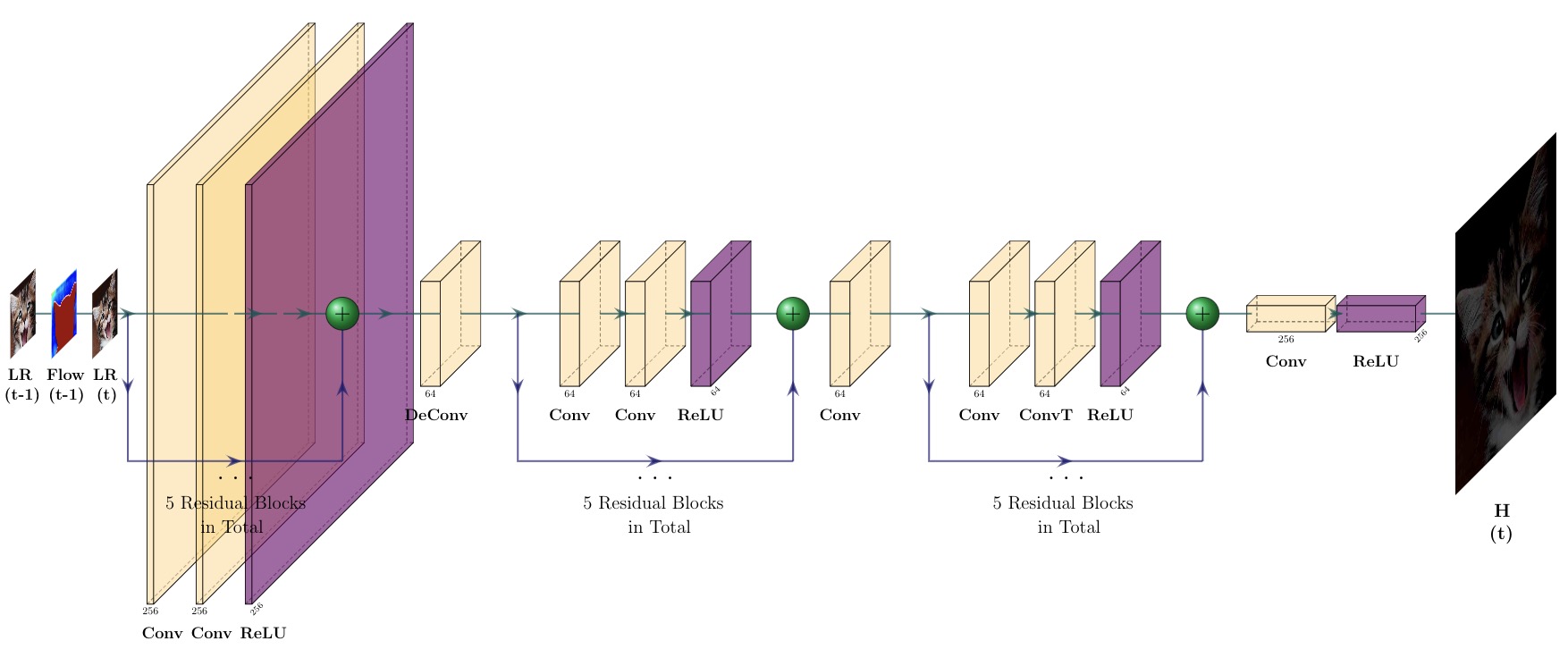} \vspace {-1mm}
    \renewcommand{\thefigure}{4}
\caption{ResNet architecture for MISR that is composed of three tiles of five blocks where each block consists of two convolutional layers with 3 $\times$ 3 kernels, a stride of 1 and padding of 1. The network uses Parametric ReLUs \cite{he2015delving} for its activations.}
    \label{fig:4}
\end{figure*}

Fig. \ref{fig:2} shows the iSeeBetter architecture that consists of RBPN \cite{haris2019recurrent} and SRGAN \cite{ledig2017photo} as its generator and discriminator respectively. Tab. \ref{tab:2} shows our notational convention. RBPN has two approaches that extract missing details from different sources: SISR and MISR. Fig. \ref{fig:3} shows the horizontal flow (represented by blue arrows in Fig. \ref{fig:2}) that enlarges $LR_{t}$ using SISR. Fig. \ref{fig:4} shows the vertical flow (represented by red arrows in Fig. \ref{fig:2}) which is based on MISR that computes residual features from (i) a pair of $LR_{t}$ and its neighboring frames ($LR_{t-1}$, ..., $LR_{t-n}$) coupled with, (ii) the pre-computed dense motion flow maps ($F_{t-1}$, ..., $F_{t-n}$). 

\begin{table}[hb]
    \caption{Adopted notation}
    \centering    
    \label{tab:2}
    \setlength{\tabcolsep}{4.5pt}
\small
    \begin{tabular}{cc}
        \toprule
$HR_{t}$ & input high resolution image\\
$LR_{t}$ & low resolution image (derived from $HR_{t}$)\\
$F_{t}$ & optical flow output\\
$H_{t-1}$ & residual features extracted from ($LR_{t-1}$, $F_{t-1}$, $LR_{t}$)\\
$SR_{t}$ & estimated HR output\\
        \bottomrule
    \end{tabular}
\end{table}

\begin{figure*}
    \vspace {-4mm}
    \centering
    \includegraphics[width=1.0\linewidth]{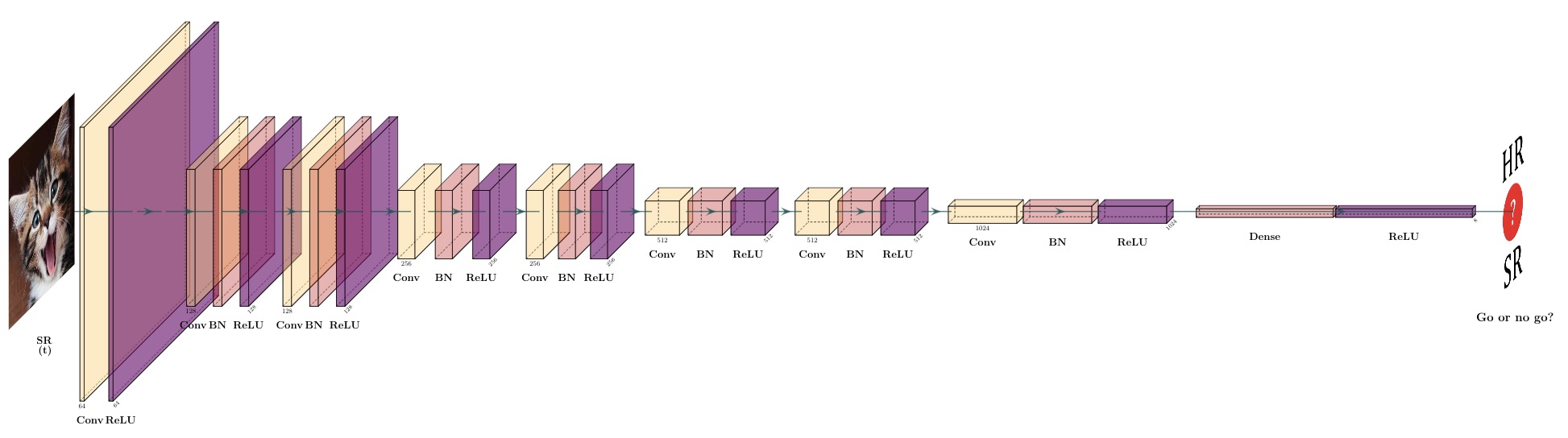} \vspace {-1mm}
    \renewcommand{\thefigure}{5}
    \caption{Discriminator Architecture from SRGAN \cite{ledig2017photo}. The discriminator uses Leaky ReLUs for computing its activations.}
    \label{fig:5}
\end{figure*}

At each projection step, RBPN observes the missing details from $LR_{t}$ and extracts residual features from neighboring frames to recover details. The convolutional layers that feed the projection modules in Fig. \ref{fig:2} thus serve as initial feature extractors. Within the projection modules, RBPN utilizes a recurrent encoder-decoder mechanism for fusing details extracted from adjacent frames in SISR and MISR and incorporates them into the estimated frame $SR_{t}$ through back-projection. The convolutional layer that operates on the concatenated output from all the projection modules is responsible for generating $SR_{t}$. Once $SR_{t}$ is synthesized, it is sent over to the discriminator (shown in Fig. \ref{fig:5}) to validate its ``authenticity''. 

\subsection{Loss functions} \label{sec:lossfunction}

\begin{figure*}[ht]
    \vspace {-4mm}
    \centering
    \includegraphics[width=0.8\linewidth]{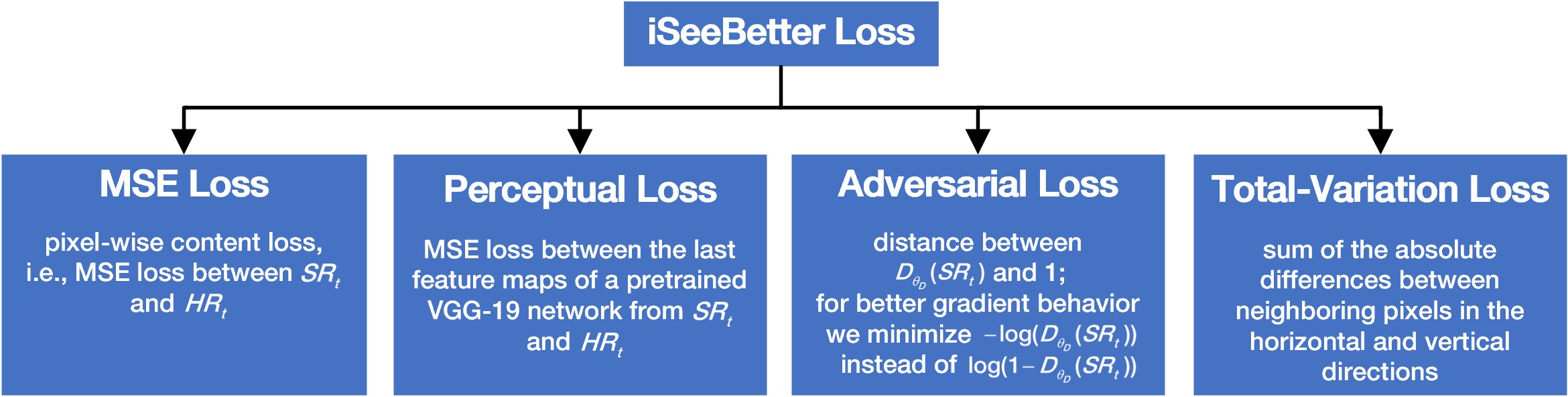} \vspace {-1mm}
    \renewcommand{\thefigure}{6}
    \caption{The MSE, perceptual, adversarial, and TV loss components of the iSeeBetter loss function.
    }\label{fig:6}
\end{figure*}

The perceptual image quality of the resulting SR image is dependent on the choice of the loss function. 
To evaluate the quality of an image, MSE is the most commonly used loss function in a wide variety of state-of-the-art SR approaches, which aims to improve the PSNR of an image \cite{hore2010image}. While optimizing MSE during training improves PSNR and SSIM, these metrics may not capture fine details in the image leading to misrepresentation of perceptual quality \cite{ledig2017photo}. The ability of MSE to capture intricate texture details based on pixel-wise frame differences is very limited, and can cause the resulting video frames to be overly-smooth \cite{cheng2012fast}. In a series of experiments, it was found that even manually distorted images had an MSE score comparable to the original image \cite{wang2002universal}. To address this, iSeeBetter uses a four-fold (MSE, perceptual, adversarial, and TV) loss instead of solely relying on pixel-wise MSE loss. We weigh these losses together as a final evaluation standard for training iSeeBetter, thus taking into account both pixel-wise similarities and high-level features.
Fig. \ref{fig:6} shows the individual components of the iSeeBetter loss function.

\subsubsection{MSE loss}
We use pixel-wise MSE loss (also called content loss \cite{ledig2017photo}) for the estimated frame $SR_{t}$ against the ground truth $HR_{t}$. 
\begin{equation}
\hspace*{-0.75em}
MSE_t = \frac{1}{{WH}}\sum\limits_{x = 0}^W {\sum\limits_{y = 0}^H {{{({{(H{R_t})}_{x,y}} - {G_{{\theta _G}}}{{(L{R_t})}_{x,y}})}^2}} }
\end{equation}
where,
${G_{{\theta _G}}}{\left( {L{R_t}} \right)}$ is the estimated frame $SR_{t}$. $W$ and $H$ represent the width and height of the frames respectively.

\subsubsection{Perceptual loss}
\cite{gatys2015texture, bruna2016super-resolution} introduced a new loss function called perceptual loss, also used in \cite{johnson2016perceptual, ledig2017photo}, which focuses on perceptual similarity instead of similarity in pixel space. Perceptual loss relies on features extracted from the activation layers of the pre-trained VGG-19 network in \cite{simonyan2014very}, instead of low-level pixel-wise error measures. We define perceptual loss as the euclidean distance between the feature representations of the estimated SR image ${G_{{\theta _G}}}{\left( {L{R_t}} \right)}$ and the ground truth $HR_{t}$.
\begin{equation}
\hspace*{-0.5em}
\begin{array}{l}
PerceptualLoss_t = \\
\frac{1}{{{W_{i,j}}{H_{i,j}}}}\sum\limits_{x = 1}^{{W_{i,j}}} {\sum\limits_{y = 1}^{{H_{i,j}}} {{{\left( \begin{array}{l}
VG{G_{i,j}}{\left( {H{R_t}} \right)_{x,y}} - \\
VG{G_{i,j}}{({G_{{\theta _G}}}(L{R_t}))_{x,y}}
\end{array} \right)}^2}} }
\end{array}\
\end{equation}
where, 
$VGG_{i,j}$ denotes the feature map obtained by the $j^{th}$ convolution (after activation) before the $i^{th}$ maxpooling layer in the VGG-19 network. $W_{i,j}$ and $H_{i,j}$ are the dimensions of the respective feature maps in the VGG-19 network.
\begin{table*}[b]
\caption{PSNR/SSIM evaluation of state-of-the-art VSR algorithms using Vid4 and Vimeo90K for 4$\times$ upscaling. Bold numbers indicate best performance.}
    \label{tab:3}
\begin{adjustbox}{width=\textwidth,center}    
    \begin{tabular}
    {ccccccccccc}
        \toprule
Dataset & Clip Name & Flow & Bicubic & DBPN \cite{haris2018deep} & B$_{\text{123}}$ + T \cite{liu2017robust} & DRDVSR \cite{tao2017detail} & FRVSR \cite{sajjadi2018frame} & RBPN/6-PF \cite{haris2019recurrent} & VSR-DUF \cite{jo2018deep} & \textbf{iSeeBetter} \\\hline
 \multirow{4}{*}{Vid4} & Calendar & 1.14 & 19.82/0.554 & 22.19/0.714 & 21.66/0.704 & 22.18/0.746 & - & 23.99/0.807 & 24.09/0.813 & \textbf{24.13/0.817} \\ 
 & City & 1.63 & 24.93/0.586 & 26.01/0.684 & 26.45/0.720 & 26.98/0.755 & - & 27.73/0.803 & 28.26/0.833 & \textbf{28.34/0.841} \\ 
 & Foliage & 1.48 & 23.42/0.575 & 24.67/0.662 & 24.98/0.698 & 25.42/0.720 & - & 26.22/0.757 & 26.38/0.771 & \textbf{26.57/0.773} \\ 
 & Walk & 1.44 & 26.03/0.802 & 28.61/0.870 & 28.26/0.859 & 28.92/0.875 & - & \textbf{30.70/0.909} & 30.50/0.912 & 30.68/0.908 \\ \hline
Average & & 1.42 & 23.53/0.629 & 25.37/0.737 & 25.34/0.745 & 25.88/0.774 & 26.69/0.822 & 27.12/0.818 & 27.31/0.832 & \textbf{27.43/0.835} \\
 \hline\hline
 Vimeo90K & Fast Motion & 8.30 & 34.05/0.902 & 37.46/0.944 & - & - & - & 40.03/0.960 & 37.49/0.949 & \textbf{40.17/0.971} \\
        \bottomrule
    \end{tabular}
\end{adjustbox}    
\end{table*}
\subsubsection{Adversarial loss}
We use the generative component of iSeeBetter as the adversarial loss to limit model ``fantasy'', thus improving the ``naturality'' associated with the super-resolved image. Adversarial loss is defined as:
\begin{equation}
AdversarialLoss_t = - log ({D_{{\theta _D}}}({G_{{\theta _G}}}(L{R_t}))
\end{equation}
where,
${{D_{{\theta _D}}}({G_{{\theta _G}}}(L{R_t})})$ is the discriminator's output probability that the reconstructed image ${G_{{\theta _G}}}{\left( {L{R_t}} \right)}$ is a real HR image. We minimize $- log ({D_{{\theta _D}}}({G_{{\theta _G}}}(L{R_t}))$ instead of $log(1 - {D_{{\theta _D}}}({G_{{\theta _G}}}(L{R_t}))$ for better gradient behavior \cite{goodfellow2014generative}.

\subsubsection{Total-Variation loss}
TV loss was introduced as a loss function in the domain of SR by \cite{aly2005image}. It is defined as the sum of the absolute differences between neighboring pixels in the horizontal and vertical directions \cite{wang2020deep}. Since TV loss measures noise in the input, minimizing it as part of our overall loss objective helps de-noise the output SR image and thus encourages spatial smoothness. TV loss is defined as follows:
\begin{equation}
\begin{array}{*{20}{l}}
{TVLos{s_t} = }\\
{\frac{1}{{WH}}\sum\limits_{i = 0}^W {\sum\limits_{j = 0}^H {\sqrt {\begin{array}{*{20}{l}}
{{{({G_{{\theta _G}}}{{(L{R_t})}_{i,j + 1,k}} - {G_{{\theta _G}}}{{(L{R_t})}_{i,j,k}})}^2} + }\\
{{{({G_{{\theta _G}}}{{(L{R_t})}_{i + 1,j,k}} - {G_{{\theta _G}}}{{(L{R_t})}_{i,j,k}})}^2}}
\end{array}} } } }
\end{array}
\end{equation}
\subsubsection{Loss formulation}
We define our overall loss objective for each frame as the weighted sum of the MSE, adversarial, perceptual, and TV loss components:

\begin{equation}
Loss{_G}_{{{_\theta }_{_{_G}}}}(SR_t)=\begin{array}{*{20}{l}}
{\rm{ }}\,\,\,\,\,\alpha{\rm{ }} \times {\rm{ }}MSE\left( {SR_t, HR_t} \right)\\
{+ {\rm{ }}\,\beta {\rm{ }} \times {\rm{ }}PerceptualLoss\left({SR_t,HR_t}\right)}\\
{+ \,\gamma \times {\rm{ }} {\rm{ }}AdversarialLoss\left( {SR_t} \right)}\\
{+\, \delta {\rm{ }}\, \times {\rm{ }}TVLoss\left( {SR_t,{\rm{ }}HR_t} \right)}
\end{array}
\end{equation}
where, $\alpha$, $\beta$, $\gamma$, $\delta$ are weights set as 1, 6\,$\times$\,10$^{-3}$, 10$^{-3}$ and 2\,$\times$\,10$^{-8}$\,respectively\,\cite{hany2019hands}. \\
The discriminator loss for each frame is as follows:
\begin{equation}
Los{s_D}_{{{_\theta }_{_{_D}}}}(SR_t) = {\rm{ }}1 - {D_\theta }_{_D}(HR_t) + {D_\theta }_{_D}(SR_t)\\
\end{equation}
The total loss of an input sample is the average loss of all frames.
\begin{equation}
\begin{array}{l}
Los{s_G}_{{{_\theta }_{_{_G}}}} = {\rm{ }}\frac{1}{N}\sum\limits_{t = 1}^N {(Los{s_G}_{{{_\theta }_{_{_G}}}}(S{R_t}))} \\
Los{s_D}_{{{_\theta }_{_{_D}}}} = {\rm{ }}\frac{1}{N}\sum\limits_{t = 1}^N {(Los{s_D}_{{{_\theta }_{_{_D}}}}(S{R_t}))} 
\end{array}\
\end{equation}

\section{Experimental evaluation}\label{sec:evaluation}

To train the model, we used an Amazon EC2 P3.2xLarge instance with an NVIDIA Tesla V100 GPU with 16GB VRAM, 8 vCPUs and 64GB of host memory. 
We used the hyperparameters from RBPN and SRGAN. Tab. \ref{tab:3} compares iSeeBetter with six state-of-the-art VSR algorithms: DBPN \cite{haris2018deep}, B$_{123}$ + T \cite{liu2017robust}, DRDVSR \cite{tao2017detail}, FRVSR \cite{sajjadi2018frame}, VSR-DUF \cite{jo2018deep} and RBPN/6-PF \cite{haris2019recurrent}. Tab. \ref{tab:4} offers a visual analysis of VSR-DUF and iSeeBetter. Tab. \ref{tab:5} shows ablation studies to assess the impact of using a generator-discriminator architecture and the four-fold loss as design decisions.

\begin{table*}[hb]
    \caption{Visually inspecting examples from Vid4, SPMCS and Vimeo-90k comparing VSR-DUF and iSeeBetter. We chose VSR-DUF for comparison because it was the state-of-the-art at the time of publication. Top row: fine-grained textual features that help with readability; middle row: intricate high-frequency image details; bottom row: camera panning motion.}
    \label{tab:4}
        \begin{center}
    \begin{tabular}{ccccc}
    \toprule
        Dataset & Clip Name & VSR-DUF \cite{jo2018deep} & \textbf{iSeeBetter} & Ground Truth \\ 
        \hline
        \noalign{\vskip 2mm}
Vid4 & Calendar & \adjustbox{valign=m}{\resizebox{115pt}{!}{\includegraphics{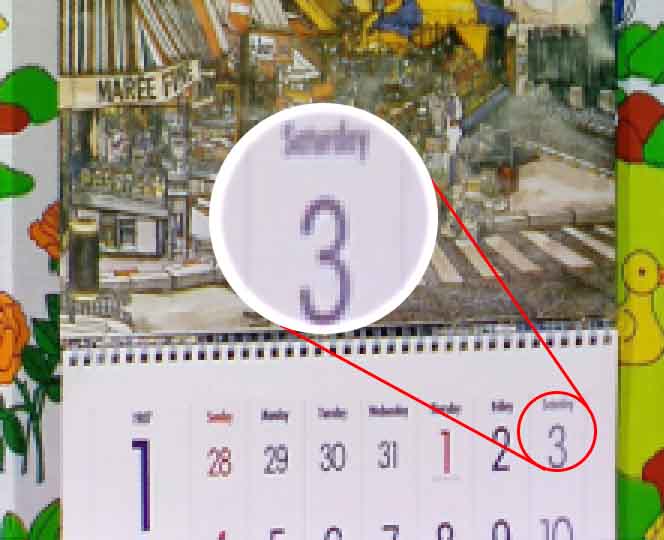}}} & \adjustbox{valign=m}{\resizebox{115pt}{!}{\includegraphics{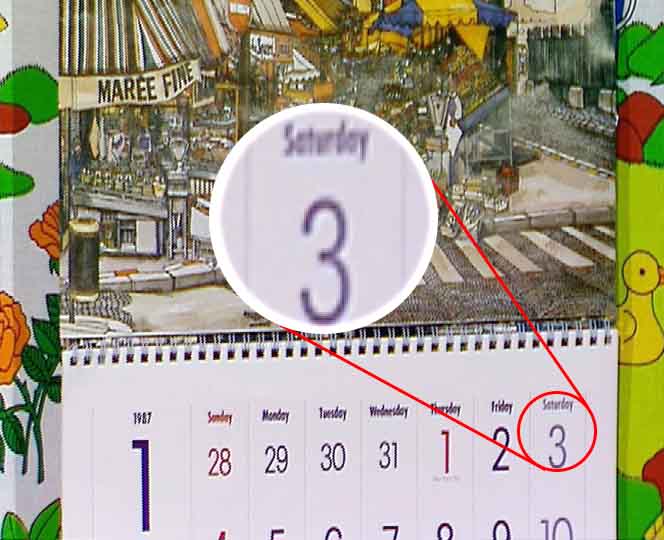}}} & \adjustbox{valign=m}{\resizebox{115pt}{!}{\includegraphics{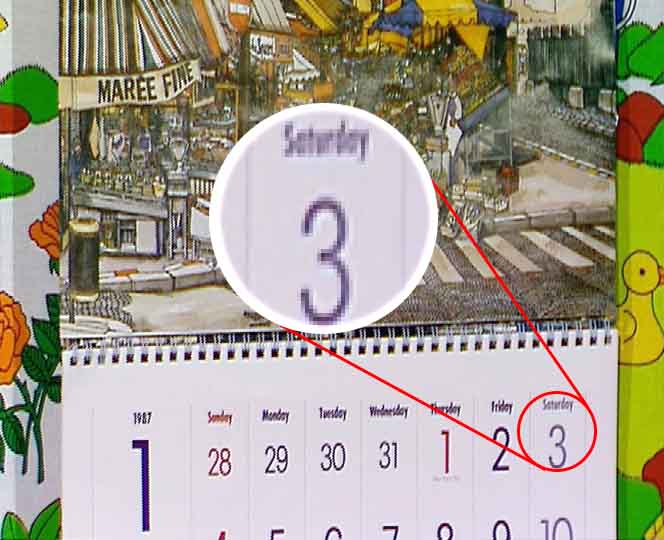}}}\\
        \noalign{\vskip 2mm}
SPMCS & Pagoda & \adjustbox{valign=m}{\resizebox{115pt}{!}{\includegraphics{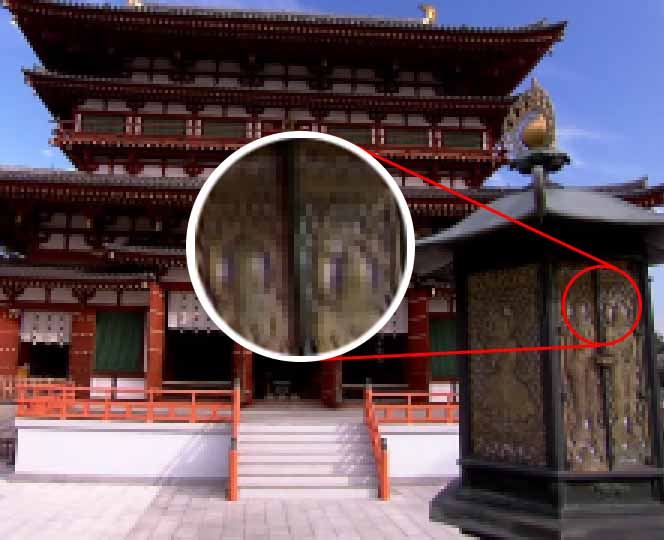}}} & \adjustbox{valign=m}{\resizebox{115pt}{!}{\includegraphics{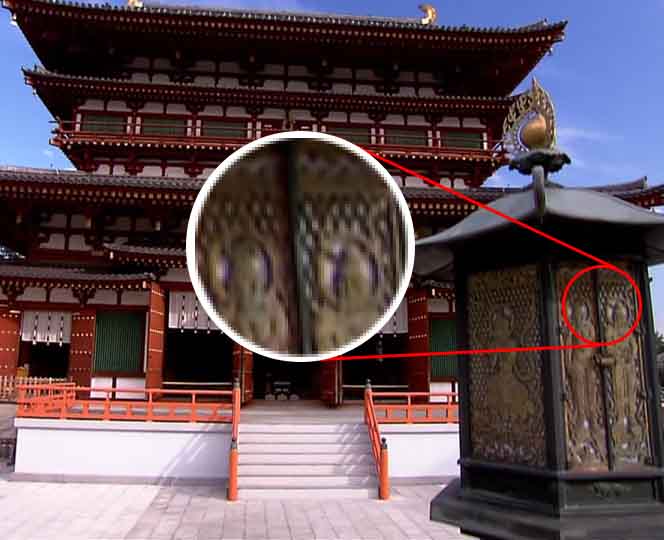}}} & \adjustbox{valign=m}{\resizebox{115pt}{!}{\includegraphics{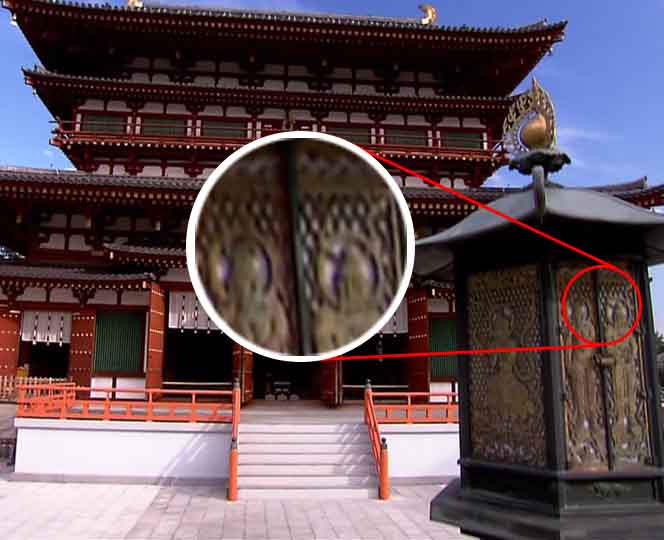}}}\\
        \noalign{\vskip 2mm}
Vimeo90K & Motion & \adjustbox{valign=m}{\resizebox{115pt}{!}{\includegraphics{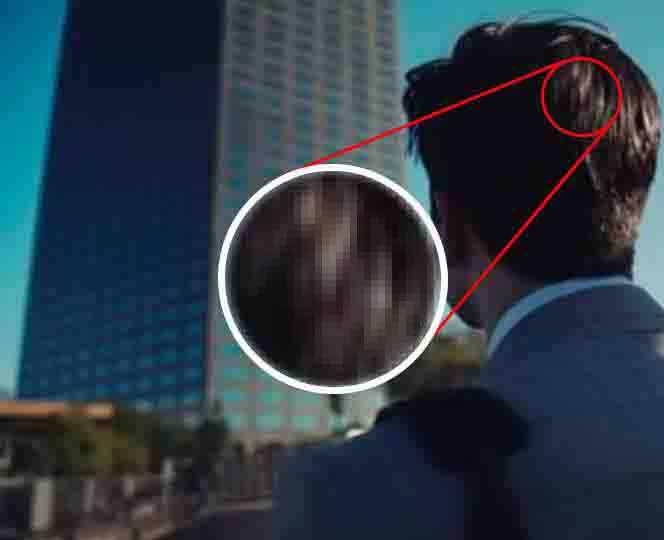}}} & \adjustbox{valign=m}{\resizebox{115pt}{!}{\includegraphics{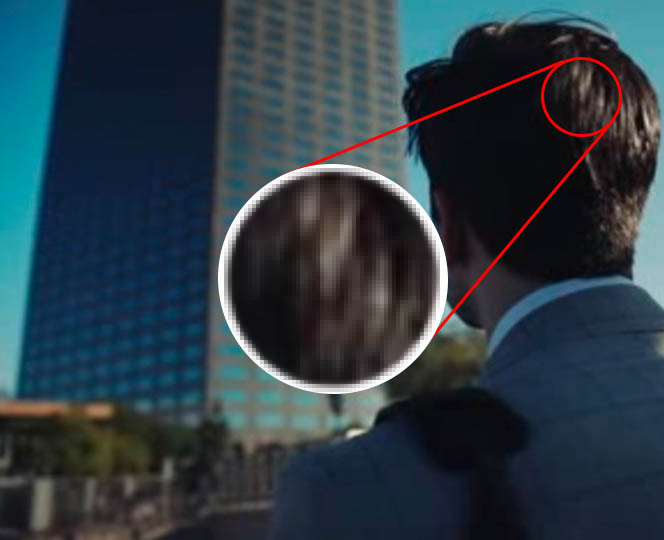}}} & \adjustbox{valign=m}{\resizebox{115pt}{!}{\includegraphics{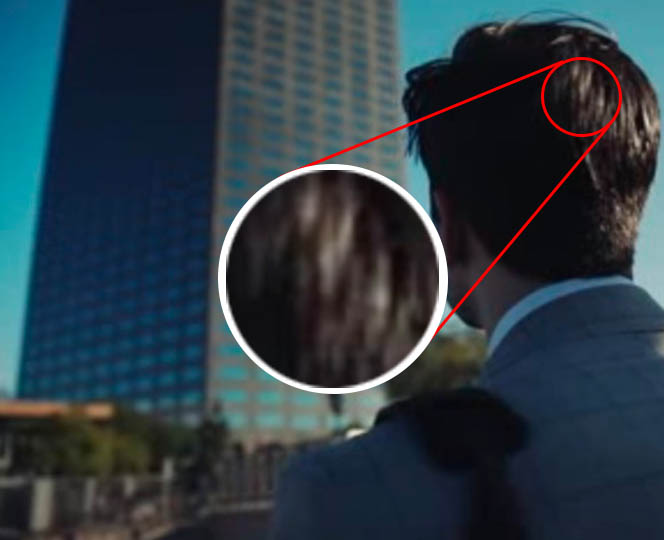}}}\\
        \noalign{\vskip 1mm}
        \bottomrule
    \end{tabular}
        \end{center}
\end{table*}

\begin{table*}[hb]
\caption{Ablation analysis for iSeeBetter using the ``City'' clip from Vid4.}
\begin{adjustbox}{width=0.9\textwidth,center}
    \begin{tabular}
    {lcc}
        \toprule
iSeeBetter Config & PSNR\\
\hline
RBPN baseline with L1 loss & 27.73 \\
RBPN baseline with MSE loss & 27.77 \\
RBPN generator + SRGAN discriminator with adversarial loss & 28.08 \\
RBPN generator + SRGAN discriminator with adversarial + MSE loss & 28.12 \\
RBPN generator + SRGAN discriminator with adversarial + MSE + perceptual loss & 28.27 \\
RBPN generator + SRGAN discriminator with adversarial + MSE + perceptual + TV loss & 28.34 \\
        \bottomrule
    \end{tabular}
\label{tab:5}
\end{adjustbox}
\end{table*}



\section{Conclusions and future work}\label{sec:conclusions}
We proposed iSeeBetter, a novel spatio-temporal approach to VSR that uses recurrent-generative back-projection networks. iSeeBetter couples the virtues of RBPN and SRGAN. RBPN enables iSeeBetter to generate superior SR images by combining spatial and temporal information from the input and neighboring frames. In addition, SRGAN's discriminator architecture fosters generation of photo-realistic frames. We used a four-fold loss function that emphasizes perceptual quality. Furthermore, we proposed a new evaluation protocol for video SR by collating diverse datasets. With extensive experiments, we assessed the role played by various design choices in the ultimate performance of iSeeBetter, and demonstrated that on a vast majority of test video sequences, iSeeBetter advances the state-of-the-art.

To improve iSeeBetter, a couple of ideas could be explored. In visual imagery the foreground recieves much more attention than the background since it typically includes subjects such as humans. To improve perceptual quality, we can segment the foreground and background, and make iSeeBetter perform ``adaptive VSR'' by utilizing different policies for the foreground and background. For instance, we could adopt a wider span of the number of frames to extract details from for the foreground compared to the background. 
Another idea is to decompose a video sequence into scenes on the basis of frame-similarity and make iSeeBetter assign weights to adjacent frames based on which scene they belong to. Adjacent frames from a different scene can be weighed lower compared to frames from the same scene, thereby making iSeeBetter focus on extracting details from frames within the same scene -- \`a la the concept of attention applied to VSR. 
\CvmAck{The authors would like to thank Andrew Ng's lab at Stanford University for their guidance on this project. In particular, the authors express their gratitude to Mohamed El-Geish for the idea-inducing brainstorming sessions throughout the project.}

\bibliographystyle{CVM}

{\normalsize  \bibliography{iSeeBetter}}

\Author{ACAD}{Aman Chadha}
{has held positions at some of the world's leading semiconductor/product companies. He is currently based out of Cupertino (Silicon Valley), California and is currently pursuing his graduate studies in Artificial Intelligence from Stanford University. He has published in prestigious international journals and conferences, and has authored two books. His publications have garnered about 200 citations. He currently serves on the editorial boards of several international journals including IJATCA, IJLTET, IJCET, IJEACS and IJRTER. He has served as a reviewer for IJEST, IJCST, IJCSEIT and JESTEC. Aman graduated with an M.S. from the University of Wisconsin-Madison with an outstanding graduate student award in 2014 and his B.E. with distinction from the University of Mumbai in 2012. His research interests include Computer Vision (particularly, Pattern Recognition), Artificial Intelligence, Machine Learning and Computer Architecture. Aman has 18 publications to his credit.}

\Author{JBRITTO}{John Britto}
{is pursuing his M.S. in Computer Science from the University of Massachusetts, Amherst. He completed his B.E. in Computer Engineering from the University of Mumbai in 2018. His research interests lie in Machine Learning, Natural Language Processing, and Artificial Intelligence.}

\Author{MMROJA}{M. Mani Roja}
{is a full professor in the Electronics and Telecommunication Department at the University of Mumbai since the past 30 years. She received her Ph.D. in Electronics and Telecommunication Engineering from Sant Gadge Baba Amravati University and her Masters in Electronics and Telecommunication Engineering from the University of Mumbai. She has collaborated across the years in the fields of Image Processing, Speech Processing, and Biometric recognition.}

\end{document}